\pdfoutput=1

\documentclass[11pt]{article}
\usepackage{EMNLP2023}
\usepackage{appendix}

\usepackage{times}
\usepackage{latexsym}
\usepackage{graphicx}
\usepackage{subcaption}
\usepackage{array}
\usepackage{amsmath}

\usepackage[T1]{fontenc}

\usepackage[utf8]{inputenc}

\usepackage{microtype}

\usepackage{inconsolata}

\title{\textit{Not wacky vs. definitely wacky: A study of scalar adverbs in pretrained language models.}}

\author{Isabelle Lorge \\
  Department of Engineering \\
  University of Oxford \\
  \texttt{engs2437@ox.ac.uk} \\\And
  Janet B. Pierrehumbert \\
  Department of Engineering \\
  University of Oxford \\
  \texttt{janet.pierrehumbert@oerc.ox.ac.uk} \\}

\begin{document}
\maketitle
\begin{abstract}
Vector-space models of word meaning all assume that words occurring in similar contexts have similar meanings.  Words that are similar in their topical associations but differ in their logical force tend to emerge as semantically close -- creating well-known challenges for NLP applications that involve logical reasoning. Pretrained language models such as BERT, RoBERTa, GPT-2, and GPT-3 hold the promise of performing better on logical tasks than classic static word embeddings. However, reports are mixed about their success. Here, we advance this discussion through a systematic study of scalar adverbs, an under-explored class of words with strong logical force. Using three different tasks involving both naturalistic social media data and constructed examples, we investigate the extent to which BERT, RoBERTa, GPT-2 and GPT-3 exhibit knowledge of these common words. We ask: 1) Do the models distinguish amongst the three semantic categories of MODALITY, FREQUENCY and DEGREE? 2) Do they have implicit representations of full scales from maximally negative to maximally positive? 3) How do word frequency and contextual factors impact model performance? We find that despite capturing some aspects of logical meaning, the models still have obvious shortfalls. 

\end{abstract}

\section{Introduction}\label{introduction}

Large pretrained language models such as BERT \citep{devlin2018bert} all rely on the assumption that the meanings of words are revealed by the company they keep \cite{harris1954distributional}; Masked Language Modelling (MLM) is a useful training objective because words in the context of the masked word provide cues to its identity. This assumption is highly appropriate for nouns and named entities. Different topics of discussion involve different entities, and in discussing any given topic, the associated entities will be referred to repeatedly. However, 
the assumption holds less well for some other classes of words that can be used in discussing virtually any topic. These include quantifiers (e.g. \textit{few, many, all}), words expressing negation (e.g. \textit{not, no, never}) and the focus of the present paper, which is scalar adverbs, such as \textit{perhaps, certainly, never, often, very, completely}. Many scalar adverbs tend to occur in similar contexts but have distinct meanings. As \citet{abrusan2018lexical} put it, distributional models tend to be `blind' to logical meanings because the latter are topic independent and thus their meanings tend to not be reflected in their distributional contexts. 


Accurate processing of scalar adverbs is pertinent to a wide variety of NLP applications, including sentiment analysis \citep{de2013good, ruppenhofer2014comparing}, entailment inferences \citep{mcnally2017scalar}, detecting contradictions, and indirect Question Answering \citep{de2010good}. BERT family models succeed without fine-tuning on a remarkable variety of tasks, 
and unsupervised embeddings from BERT base have been used fairly successfully to rank gradable adjectives according to their scalar position on half-scales from a neutral to an extreme value (e.g., \textit{warm < hot < scalding}) \citep{soler2020bert}. This result suggests that the BERT architecture might also be successful in exploiting diffuse or indirect cues for the selection of scalar adverbs. However, the same models have had little success in representing negation \citep{ettinger2020bert} or antonymy \citep{talmor2020olmpics}. In fact, even large language models (LLMs) still struggle substantially with negation \cite{truong-etal-2023-language}. \citet{soler2020bert} do not evaluate adjectival antonyms, such as \{\textit{cold, hot}\}. However, it is worth noting that scalar adverbs intrinsically include antonym pairs, (e.g.,\{\textit{never, always}\}). Furthermore, the word \textit{not} is syntactically possible wherever a scalar adverb can occur, yet would express a contradictory meaning to any positive scalar adverb. 

These latter observations suggest that scalar adverbs might present important challenges for LLMs, and point to the need for a deep assessment. We look at full scales, and directly compare scalar adverbs to explicit negation. We consider the 24 adverbs in Table 1, selected through the process described in Section 3. 
These vary widely in frequency, with \textit{very} found 22818 times in the Reddit slice and \textit{frequently} found only 52 times. At 40986 occurrences, \textit{not} is more frequent than any scalar adverb (see Appendix for full adverb frequencies). 

\begin{table}
\begin{tabular}{|m{2.2cm}|m{4.5cm}|} 
 \hline
 \textbf{Category} & \textbf{Adverbs}  \\ 
\hline
 \textbf{MODALITY} (14.8\%) & \textit{\{maybe,  perhaps,
possibly\}, arguably, probably,
 actually,
 certainly, definitely} \\ 
\hline
\textbf{FREQUENCY} (5.3\%) &   \textit{never, occasionally, sometimes, often, generally, usually, frequently, always}\\
\hline
\textbf{DEGREE} (46.8\%) & \textit{hardly, slightly, basically, pretty, quite, very, really,  completely}  \\
\hline
\end{tabular}
\caption{Scalar adverbs in each semantic category ranked by scalar position using semantic theory and Word{Net} definitions. Bracketed items are tied. Percentages for each category are the overall percentage of that category in the Reddit slice in relation to the set containing the target adverbs and \textit{not}.}
\label{table:1}
\end{table}

We ask the following questions: 1) Are pretrained language models able to distinguish between different semantic categories of scalar adverbs? 2) Do they have general representations of full scales for adverbs, from maximally negative to maximally positive? 3) Do their representations support success in three tasks: ranking, MLM, and evaluating entailments? 4) In what way do the patterns of success and failure relate to word frequency and contextual factors?

\section{Background}\label{background}
We begin by introducing some concepts from linguistic semantics and pragmatics that motivate our study. 

\subsection{Scales and operators}
The workhorse of document retrieval is the fact that the topic under discussion hugely influences what entities people refer to. According to semantic theory, individual unique entities are referred by proper nouns; common nouns refer to sets of entities.  The bursts in uses of proper and common nouns associated with the current topic provide the basis for the distributional hypothesis about word meanings. 

In contrast to nouns, many other types of words have more complicated semantic structures. \citet{partee1992syntactic} develops a typology of word types according what implicit variables they contain. In 
\citet{altmann2009beyond}, this typology is simplified and applied to explain why some types of words are typically much less bursty than nouns. Of particular relevance here is the distinction between entities and operators. Operators are words that have hidden variables in their semantic representations, which are supplied by the context. The many different ways of filling in these hidden variables means that they can be used in many different contexts. As a corollary, \citet{altmann2009beyond} demonstrate that they are much less bursty than words referring to entities.   

One much-studied class of operators is gradable adjectives such as \textit{hot} or \textit{tall}. These  position the expression they modify on a scale. By using them, the speaker indicates that the modified expression has a position on the scale that is more extreme than a given threshold, which is inferred from the context \citep{lassiter2013context}. 
For example, by hearing someone described as \textit{tall}, or water described as \textit{hot}, the listener will apply their knowledge about people's heights, or water temperatures,  
to infer that the height or temperature being described is significantly above its typical value.  This means that formal semantic representations of \textit{tall, hot} each contain a hidden variable, representing the threshold whose value is contextually determined. For negative adjectives such as \textit{short, cold}, the corresponding inference is that the value falls below a critical threshold. 


The adverbs in our study themselves modify scalar adjectives, introducing a further level of abstraction.  Consider the following sentences:

\begin{enumerate}
    \item \textit{The water is \textbf{VERY} hot} (DEGREE)
    \item \textit{The water is \textbf{OFTEN} hot}. (FREQUENCY)
    \item \textit{The water is \textbf{PROBABLY} hot}. (MODALITY)
\end{enumerate}

 Adverbs of DEGREE simply move the degree threshold of the original adjective \citep{bennett2018extremely}, i.e., \textit{very hot water} has an inferred range of temperatures higher than \textit{hot water}. Adverbs of FREQUENCY, on the other hand, do not act on the (continuous) intensity of a single event, but rather describe a point on a scale of discrete occurrences of the relevant property \citep{doetjes2007adverbs}.  Lastly, modal or epistemic adverbs do not modify the adjectival property itself, but instead are an evaluation of the likelihood of the property by the speaker \citep{lang1979status}. Because they pertain to different scales, these categories can be freely combined without contradiction;  e.g a person may be \textit{often slightly angry}, \textit{certainly slightly angry}, \textit{occasionally very angry}, or \textit{sometimes definitely angry}.

Thanks to their hidden, contextually determined variables, operators are freely available in a wide variety of contexts. It follows that the context provides little information about which operator was selected in any instance. While  \textit{always} and \textit{never} differ greatly in their logical force, they may differ little in their contexts of use. It follows that the particular context may provide little information about which one the speaker actually selected. 


\subsection{Entailment}

One of the main tasks used to evaluate natural language inference is an entailment task. We accordingly define an entailment task to probe how well LLMs reason about scalar adverbs. For a typical entailment task, (e.g., MNLI, \citealp{N18-1101}) crowdworkers are asked to label the relations between two ordered sentences as entailment, neutral or contradiction. Thus the NLP literature considers sentence A to entail sentence B if B can be normally inferred from A. This is a loose definition in relation to the research literature in linguistics, which has since \citet{grice} drawn a critical distinction between entailment (a semantic concept based on the logical meanings of words) and implicature \citep{huang}. Implicatures are inferences made in the context of the discourse, including relevant real world knowledge, which do not meet the strict criteria for logical entailment. Unlike entailments they are readily cancelled without engendering contradictions. For example, the MNLI dataset characterizes \textit{... people began to form a line ... } as ``entailing'' \textit{... people formed a line ...}. However, initiating an action does not necessarily mean that the action was completed; something might happen to prevent this.   Pervasive confusion between entailment and implicature in the NLP literature means that levels of success on entailment tasks may be inflated due by common associations of events, rather than logical reasoning. 

Here, we confine our attention to entailment in the strict logical sense. 
A critical observation is that entailment may only be strictly defined over word relationships that involve the same scale. For example,  if it is \textbf{very} cold, it is at least \textbf{somewhat} cold but if it is \textbf{very} cold, it is unclear whether it is at least \textbf{often} cold. Previous work indicates that pretrained language models struggle with entailment relations, such as hypernymy and hyponymy \cite{guerin-chemla-2023-bird}. 

\section{Materials}\label{materials}

Building on the approach of \citet{ribeiro-etal-2020-beyond} and \citet{rottger-etal-2021-hatecheck}, we had the goal of designing a balanced diagnostic dataset for probing how well models capture the meanings of scalar adverbs. Our primary dataset consists of 960 items, which are based on posts from the year 2015 in the Reddit politosphere dataset introduced in \citet{hofmann2022reddit}. This slice represents about 6GB of data from a range of political subreddits (e.g., \textit{r/conservative} or \textit{r/anarchist}). It offers naturalness and domain consistency as well as a certain amount of diversity in linguistic usage. 

To select the posts, we first used Spa{C}y \cite{spacy2} to extract phrases of the form `\textit{ADV ADJ.}' where there is a dependency between the adjective and the adverb. We take only phrases in which this construction occurs in final position, so that the phrases are also guaranteed to be well-formed in isolation. We then include the previous context from the same post, up to a maximum of 40 words, cutting at a sentence boundary. In semantic theory, preceding context is known to be important \cite{beaver2001presupposition, roberts1995domain}, but this factor is unfortunately disregarded in most of the related NLP research. Aiming for at least 40 different adjectives to occur with each target adverb, we selected 8 distinct adverbs that expressed the speaker's judgment on a scale of likelihood (MODALITY), 8 that express a position on a temporal scale (FREQUENCY), and 8 with more general applicability (DEGREE).  The adverbs were selected to span the full range of each scale, and hence include adverbs with negative force i.e., \textit{hardly} and \textit{never}. To shed light on the contrast between scalar adverbs and outright negation, the word \textit{not} is reserved as a benchmark and not used as a target. Both the adverbs themselves, and the \textit{ADV ADJ} bigrams, were selected to span the range of available frequencies to the extent possible, using Google Ngram \citep{lin2012syntactic} frequencies.   The target adverbs are listed in Table \ref{table:1}.

According to \citet{paradis1998degree}, some of our chosen scalar adverbs are `maximizers' (e.g., \textit{completely}) which tend to occur with extreme adjectives (e.g., \textit{freezing}) or limit adjectives (e.g., \textit{dead}). Others typically combine with stereotypically scalar adjectives (e.g., \textit{cold}). However, these are tendencies rather than rules \citep{kamoen2011absolutely}. Indeed,  phrases such as \textit{very dead} or \textit{completely cold} can be perfectly acceptable  in some contexts (e.g., \textit{I can assure you he was very dead}).  Therefore, we do not restrict our phrases to traditional scalar adjectives and include any occurrences involving the target adverbs. Example items and targets can be seen in Table \ref{targetexamples}. The dataset features (word lengths and adjective overlap between adverbs) can be found in Table \ref{naturalisticdatasetchar} in the Appendix.

\begin{table}
\begin{tabular}{|m{2.2cm}|m{4.4cm}|} 

 \hline
 \textbf{Target} & \textbf{Context}  \\ 
\hline
\textit{certainly } & You are conflating the issue. Slavery was not moral but it was  \textit{[MASK] legal}.\\
\hline
\textit{frequently } & Doesn't really matter what republicans say, democrats are going to call them racist.  Because what Republicans say is \textit{[MASK] racist}. \\ 
\hline
\textit{very}  &  You need verifiable proof.  I mean, it's not like saying you're self trained is \textit{[MASK] reputable}.  \\
\hline
\end{tabular}
\caption{Example target phrases and sentences for MLM task}
\label{targetexamples}
\end{table}

These items were used as such in an MLM task. The same target adverbs are also used in the entailment task, but constructing items with templates rather than using the natural contexts. For the adverb ranking task, we combined each target scalar adverb with (the same) 40 common adjectives to try and get an average of generic embeddings: \textit{able, bad, big, black, clear, different, early, easy, economic, federal, free, full, good, hard, high, human, important, international, large, late, little, local, low, military, national, new, old, only, political, possible, public, real, recent, right, small, social, special, strong, white, young}. We describe the construction of the entailment items below in \ref{sec:entailment}.

\section{Tasks}

Similar to \citet{talmor2020olmpics}, \citet{liu2021pre} and \citet{ jiang2022promptbert}, our main tasks are zero-shot evaluations without fine-tuning so as to examine the representations learned from pretraining. We first evaluate the extent to which the rankings in Table 1 can be recovered from the embedding space in BERT and RoBERTa. We then look at MLM (as one of the training objectives for the models) and finally entailment (as a canonical logical task). In the Appendix, we also consider a model fine-tuned on a Natural Language Inference dataset (MNLI, \citealp{N18-1101}).

\subsection{Ranking adverbs by scalar position}

Our first question is whether the rankings of the various scalar adverbs along their relevant scales are observable in the embeddings. Resources that provide scalar rankings for adverbs are scarce, and  the few available, such as \citet{taboada2011lexicon}, confound scalar position with other factors.  Therefore, we defined our own gold standard (cf. Table \ref{table:1}), on the basis of WordNet definitions.  

 We first applied both methods described in \citet{soler2020bert} for assessing the scalar position of scalar adjectives.  Their first method (SIM) uses a reference point, specifically the top end of each scale, and computes the cosine similarity for each target from the reference point; the similarity should decrease as we move down the scale.  Their second method (DIFF) uses the difference between between the maximum and minimum words on a scale to define an abstract vector of scale position; the cosine similarity of any word to this vector is taken to indicate its scale position.  

Broadly inspired by \citet{maillard-clark-2015-learning} and \citet{socher-etal-2012-semantic}'s work on semantic composition for nouns, we also devised a third method (AdjDIFF). Reasoning that the effect of the scalar adverb on the contextual embedding of the adjective may correlate with the scalar adverbs' position on the scale, we obtain embeddings for each adjective with and without the scalar adverb.  We subtract the unmodified embedding from the modified embedding of the adjective to obtain an estimate of the vector for the scalar adverb. We then take the cosine similarity of each resulting vector with the same referent vector as in the DIFF method and average them to obtain the final cosine similarity value. 

\begin{equation}
\begin{split}
        rank_{AdjDiff} = cos(\vec v_{adj(+adv)}-\vec v_{adj}, \\
        \vec v_{top} - \vec v_{end})
\end{split}
\end{equation}

The results for the AdjDiff method, which was overall the best performing, can be found in Table \ref{table:ranking}. We report the pairwise accuracy, Spearman $\rho$ and tie corrected Kendall's $\tau$ for RoBERTa, BERT large and BERT base. (See the Appendix for the results using the \citet{soler2020bert} methods). 

\begin{table*}
\centering
\begin{tabular}{|l|l|l|l|l|l|l|l|}
\hline
& \textbf{Pacc.} 
& \multicolumn{3}{c|}{\textbf{Spearman $\rho$}} 
& \multicolumn{3}{c|}{\textbf{Kendall $\tau$}} \\
\hline
 
\textbf{\textit{BERT-b}} &  0.60 &

\textit{f:} 0.68 &
\textit{m:} 0.77  &
\textit{d:} 0.32
& 
\textit{f:} 0.52 &
\textit{m:} 0.66 &
\textit{d:} 0.23
\\
\hline

\textbf{\textit{BERT-l}} & \textbf{0.64} & 

\textit{f:} \textbf{0.78} &
\textit{m:} 0.88 &
\textit{d:} 0.39
& 
\textit{f:} \textbf{0.62} &
\textit{m:} 0.77 &
\textit{d:} 0.24
\\
\hline

\textbf{\textit{ROBERTA}} & 0.53 & 

\textit{f:} -0.32 &
\textit{m:} 0.77 &
\textit{d:} 0.64
& 
\textit{f:} -0.24 &
\textit{m:} 0.67 &
\textit{d:} 0.52
\\
\hline
\end{tabular}
\caption{\label{citation-guide}
Results of scalar ranking tests \textbf{BERT-b} = BERT base; \textbf{BERT-l} = BERT large; \textbf{ROBERTA}; \textit{f} = FREQUENCY, \textit{m} = MODALITY. \textit{d} = DEGREE). 
}
\label{table:ranking}
\end{table*}

Overall, the performance is worse than what \citet{soler2020bert} obtained for adjectival half-scales. Adverb ranking may be more difficult than adjective ranking and/or full scales may be difficult to tank than half scales. However the overall accuracy of $0.64$ for the BERT-large method indicates that some information about the relations has been captured. It is interesting that BERT-large performs better than RoBERTa, for which the existence of negative values of Spearman $\rho$ and Kendall's $\tau$ is particularly problematic. We also note that the FREQUENCY category shows the worst performance.  

\subsection{Masked Language Modelling}

MLM is one of two training objectives for BERT, and the only  objective for RoBERTa. For BERT or RoBERTa to form good representations of the scalar adverbs, the larger context needs to contain information about which ones are most likely in any given instance. According, we directly evaluate the raw Masked Language Modelling outputs for the target phrases we selected. How well does each model predict a scalar adverb in a context when it is masked? 

Based on the discussion in Section \ref{background}, we expect this task to be extremely difficult. For most of our examples, it appears that humans would be unlikely to succeed in predicting the masked word. However, through learned attention weights, BERT and RoBERTa integrate information over a large time window, potentially performing better than expected. Possible sources of information about the scalar adverb include collocations or selectional restrictions involving the following adjective, and rhetorical devices or idiomatic expressions that involve the preceding context. Hence, we systematically explore the success of MLM in predicting a masked adverb.  If MLM is successful, that means that the predictive information is present in a way that is not intuitively evident, whereas if MLM fails, that tends to suggest that predictive information is simply lacking in the text stream. 

In light of the difficulty of the task, we report two measures. One is the Mean Reciprocal Rank (MRR) for the original adverb, which scores high if the adverb that occurred is ranked highly even if it is not the one that actually appeared
\footnote{Unlike \citet{truong-etal-2023-language}, we do not use Weighted Hit Rate (WHR) since this requires a fixed set of wrong predictions.}
. MRR is defined as
\begin{equation}
    MRR = \frac{1}{N}\sum_{n=1}^{N} \frac{1}{rank_{adv}}
\end{equation}

Where $N$ is the number of items for the original target adverb and $rank_{adv}$ is the rank of the original target adverb among the model's predictions. 

Our other metric is whether the model ranked the original adverb above \textit{not}. In our materials, replacement with \textit{not} is always syntactically possible, but would either contradict the previous context, or contradict what the speaker actually said.  Thus, \textit{not} should generally be ranked as less likely than \textit{any} scalar adverb, with the exception of other negative polarity items such as \textit{hardly} and \textit{never}\footnote{Since we do not have precise numbers of negation acceptability for our examples (a difficult issue in naturalistic contexts which is beyond the scope of this article, see Limitations), this metric should be regarded as indicative only, in contrast to MRR.}.  We test three models in the BERT family: BERT base, BERT large and RoBERTa large. We also test GPT-2 \cite{radford2019language}, which unlike BERT family models is an autoregressive model and helps us to see whether the right-hand context contains relevant information.  We use the pretrained cased BERT large and BERT base from the Huggingface's \textit{transformers} library \citep{DBLP:journals/corr/abs-1910-03771}, replacing our target scalar adverb with the [MASK] token and obtaining the logits which we convert back into probabilities.

We use neutralised versions of the sentences e.g., `\textit{is ADV ADJ.}' as a baseline for predictions. This provides the BERT-family models with a syntactic cue plus any selectional biases from the ADJ.

\subsection{MLM Results}

Results can be found in Table \ref{resultsreddit}. All models perform extremely poorly in the neutral context, indicating that adjectives alone are not sufficient to predict adverbs. (The GPT2-neutral condition of course has no success, since GPT2 does not use right-hand context). The results for the full context are better. Both BERT large and BERT base get a significant boost from the full context both in upranking the original adverb (MRR doubling for both models) and in ranking the original adverb above negation. RoBERTa performs best overall.  The MODALITY category gets the highest boost from context, from 0.02 to 0.57, this may be due in part to the fact that English lacks a negative item in this category. Error analysis shows that BERT still yields negation as the top prediction or among the top predictions even in cases where the context makes it unlikely (eg., \textit{not} is the top prediction for the first two examples in Table \ref{targetexamples}). 

\begin{table*}
\begin{tabular}{|m{1cm}|m{0.4cm}|m{0.4cm}|m{0.4cm}|m{0.4cm}|m{0.4cm}|m{0.4cm}|m{0.4cm}|m{0.4cm}|m{0.4cm}|m{0.4cm}|m{0.4cm}|m{0.4cm}|m{0.4cm}|m{0.4cm}|m{0.4cm}|m{0.4cm}|}
\hline
&  \multicolumn{2}{c|}{\textbf{BERTb(c)}} & \multicolumn{2}{c|}{\textbf{BERTb(n)}} & 
\multicolumn{2}{c|}{\textbf{BERTl(c)}} & 
\multicolumn{2}{c|}{\textbf{BERTl(n)}} &
\multicolumn{2}{c|}{\textbf{RoBl(c)}} & 
\multicolumn{2}{c|}{\textbf{RoBl(n)}} &
\multicolumn{2}{c|}{\textbf{GPT2(c)}} & 
\multicolumn{2}{c|}{\textbf{GPT2(n)}}  \\

\hline
& \textit{ac.} &\textit{r} & 
\textit{ac.} &\textit{r} & 
\textit{ac.} &\textit{r} & 
\textit{ac.} &\textit{r} & 
\textit{ac.} &\textit{r} & 
\textit{ac.} &\textit{r} & 
\textit{ac.} &\textit{r} & 
\textit{ac.} &\textit{r} 
\\
\hline

\textit{FREQ.} & .22 & .11 &
.02 &  .04 & 
 .36 & .15 &
.06 & .05 &
\textbf{.52} & \textbf{.21} &
.06 & .04 &
.08 & .04 &
.00 & .01
\\
 \hline
\textit{MOD.}  & .19 & .09 & 
.00 & .02 & 
.33 & .11 & 
.01  & .02 &
\textbf{.57} & \textbf{.18} &
.02 &  .02 & 
.09 & .05 &
.00 & .01 
\\
\hline

\textit{DEG.} & .4  & .2 &
.08 &   .10 & 
.53 & .24 & 
.22 &  .15 & 
\textbf{.67} & \textbf{.28} & 
.29 & .16 & 
.16 & .06 & 
.00 &  .01 
\\
\hline
\hline
\textbf{avg.} & .27 & .14 & 
.03 &  .05 &
.41 &  .17 &
.17 & .07 &
\textbf{.59} & \textbf{.22} &
.12 & .07 &
.11 & .05 &
.00 & .01
\\
\hline
\end{tabular}
\caption{\label{citation-guide}
Accuracies (ac.), i.e., number of times the original adverb was ranked above \textit{not} and Mean Reciprocal Rank (r) for each adverb and semantic category.
\textit{(c)} = full context; \textit{(n)} = neutral context.
}
\label{resultsreddit}
\end{table*}


\begin{figure*}[ht!]
        \centering
        \begin{subfigure}[b]{0.475\textwidth}
            \centering
            \includegraphics[width=\textwidth]{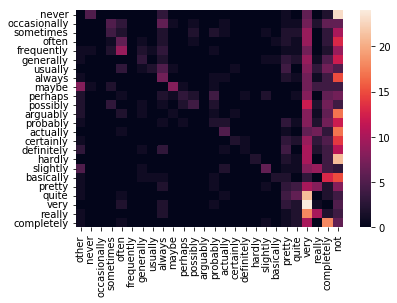}
            \caption[]%
            {{\small BERT large with context}}    
            \label{fig:mean and std of net14}
        \end{subfigure}
        \begin{subfigure}[b]{0.475\textwidth}   
            \centering 
            \includegraphics[width=\textwidth]{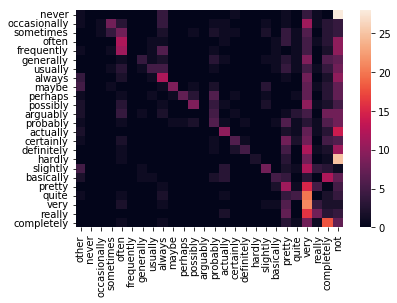}
            \caption[]%
            {{\small RoBERTa large with context}}    
            \label{fig:mean and std of net34}
        \end{subfigure}
        \caption[]
        {\small Heatmap of confusion matrices per scalar adverb for each model with context (items are grouped by semantic category). In the interest of space considerations, we only show results for BERT large and RoBERTa} 
        \label{fig:heatmaps_context}
    \end{figure*}

We construct confusion matrices between original target adverb and the top output prediction for each example for each model and context. We select the first of our target adverbs in the top 10 outputs, or the category `other' when none of the top 10 predictions appears in our list. The heatmaps with context can be seen in Figure \ref{fig:heatmaps_context} (the full set of heatmaps, including outcomes for the neutral context, can be found in Figure \ref{fig:heatmaps_accuracies} in the Appendix). While there is some indication of ability to predict the original adverb from BERT (i.e., the faintly lit up diagonal), it is clear that the decision is strongly driven by prior frequency effects, with \textit{not} and \textit{very} topping the predictions for all targets. RoBERTa gets a better performance, as is evidenced by the more strongly lit up diagonal, but frequency effects still dominate (the vertical lines for items such as \textit{very}). The figure is laid out so that within-category confusions would show up on a block diagonal pattern. No such pattern exists, indicating that scalar adverbs within the same semantic category do not emerge as particularly similar. 

The fact that the models enjoy some success when provided with the left-hand context of the target indicates that the left-hand context -- unlike the right-hand context -- contains information about which scalar adverb is more likely in which instance. However, because of the naturalistic and varied nature of our examples, it is uncertain to what extent this success derives from distributional patterns versus logical relationships. 

\subsection{Scalar entailment task}\label{sec:entailment}

The adverb rankings in Table 1 can readily be translated into entailments. Evaluating putative entailments is an established test of logical reasoning: \textit{It is always cold} entails \textit{It is sometimes cold}, but the reverse is not true. If the models have reliably learned the logical relations between scalar adverbs during pretraining, they should rank completions which create correct entailments higher than completions which create contradictions. 
We set up an entailment task as a MLM task in two conditions, which we illustrate now with example items constructed from the ADV ADJ combination \textit{often special}. For the BELOW condition, we create items where we expect an adverb which is below the premise adverb on the relevant scale, e.g. \textit{If it is often special, then it is at least [MASK] special} (sometimes, occasionally, etc.). For the ABOVE condition, we expect a completion which is above the premise adverb, e.g., \textit{If it is [MASK] special, then it is at least sometimes special} (often, usually, etc.). We craft items using eight different templates for each condition (varying order of premise and mask as well as subordinating conjunction). These can be found in the Appendix along with detailed results for each template. We omit the scalar adverbs which are technically negations (\textit{hardly, never}) and omit bottom scalar adverbs (\textit{sometimes, maybe, slightly}) for the BELOW condition and top scalar adverbs (\textit{always, definitely, completely}) for the ABOVE condition, since no correct answer is available for these. In contrast to the natural items for the MLM task,  \textit{not} is always a logically impossible completion for all items in the entailment task. 


We used 160 adjectives systematically varied in frequency. From our Reddit data, we selected adjectives in the \textit{low} frequency range (log frequency -18 to -14), \textit{medium} frequency range (log frequency -14 to -10) and \textit{high} frequency range (log frequency -10 to -6). We use the recent \textit{wordfreq} Python library for this purpose \citep{robyn_speer_2022_7199437}, which is sourced from the \textit{Exquisite Corpus} project and compiles multilingual data from 8 domains. In addition, we selected 40 pseudo words as adjectives from the highest ranked items in \citep{needle20224} under the constraint that they are not compounds of real words, so that the pretrained models' WordPiece tokenization does not introduce any previous information. We combine these 160 adjectives with our target adverbs for each template and condition to create a dataset of 40960 sentences for which we collect MLM completions from BERT large and RoBERTa.

\begin{table*}
\centering
\begin{tabular}{|c|m{1.2cm}|m{1.2cm}|m{1.6cm}|m{1.6cm}|} 
 \hline
 & \textbf{BERTl (acc.)} & \textbf{BERTl (triv.)}  
  & \textbf{RoBERTa (acc.)} & \textbf{RoBERTa (triv.)} \\


\hline
\textit{with negation} & 0.35 & 0.20   & 0.42 & 0.17   \\
\hline\textit{without negation} & \textbf{0.88} & \textbf{0.33}   & \textbf{0.88} & \textbf{0.25}  \\
\hline
\hline
\textit{BELOW} & 0.53 & 0.25   & 0.60 & 0.21  \\
\hline
\textit{ABOVE} & \textbf{0.69} & \textbf{0.28}   & \textbf{0.71} & \textbf{0.14}  \\
\hline

\end{tabular}
\caption{Results for scalar entailment dataset (BERT large and RobERTa). \textit{(without negation)} = not taking into account negations as answers, \textit{acc.} = accuracy, \textit{triv.} = number of trivial answers (e.g., \textit{If it's sometimes ADJ, it's sometimes ADJ}, which we do not take into account for accuracies), \textit{BELOW} = expects item below on the relevant scale, \textit{ABOVE} = expects item above on the relevant scale (best in bold).}
\label{entailresults}
\end{table*}


\subsection{Entailment results}

As in the MLM task, to construct confusion matrices we select the first answer on our adverbs list from the top 10 completions of the model  (including negative items: \textit{not, hardly} and \textit{never}), and the category `other' if no completion in the top 10 is found in the relevant category. To calculate accuracies, we exclude trivial answers (e.g., \textit{If it is sometimes strong, then it is sometimes strong}) as well as `other' answers which do not pertain to the target semantic category (e.g., \textit{mostly} when the category is temporal). However, we do report trivial answer percentages separately. A model which randomly picks an adverb in the relevant category produces 0.13 trivial answers. 

The results when taking into account negative answers are very poor.  The models output a high percentage of negations (especially `not') even though negations constitute logical contradictions for all sentences in the entailment dataset. 

To get a more nuanced picture of the models' behaviour and support comparisons with the MLM results, we also build heatmaps without taking negative answers into account, and calculate accuracies without negations. 
Results (with and without negations) can be found in Table \ref{entailresults}. More detailed results by adjective frequency can be found in Table \ref{entailresultsdetailed} in the Appendix. Both sets of heatmaps (including negative answers) can also be found in Figures \ref{fig:heatmaps_with_not} and \ref{fig:heatmaps_no_not} in the Appendix. 

When choosing the top relevant answer excluding negations the results improve drastically to near ceiling. However, the models most likely benefit from biases in both conditions. In the ABOVE condition, the most frequent items (\textit{always, actually, very}) constitute correct answers in a majority of cases. In the BELOW condition all templates have a textual hint (\textit{at least/at most}) which strongly suggests an item outside the top of the scale (\textit{?at least/at most always/completely/definitely)}. The benefits from the bias towards high frequency top-of-scale adverbs in the ABOVE condition are probably stronger than those from the textual hints in the BELOW condition,  which may explain why the models perform worse in the BELOW condition.

The adjective frequency has little effect on performance, which is in fact slightly better for low frequency adjectives and pseudowords.  This provides further evidence that the models are not memorizing ADV-ADJ combinations. This observation is strengthened by adverb frequency effects which prevail across scales (i.e., vertical lines in the heatmaps e.g., `slightly', `really') in both BERT large and RoBERTa. The rate of trivial answers in both models also appears to be far above what would be expected from humans (although this remains to be tested by collecting human judgments). 

To summarize, both BERT large and RoBERTa show very poor ability to distinguish between non-negative scalar adverbs and negation. The models perform well if we consider first completions excluding negations. However they most likely benefit from frequency biases and it is doubtful whether they learned a separate logical representation of the adverbs' scalar property. The models also output a high number of trivial, uninformative completions and seem affected by noise associated with frequent adjectives. Finally, differing performance on the ABOVE and BELOW conditions, which are logically equivalent, indicates that neither model has a general grasp of the underlying logic.

\section{Conclusion}
The goal of this paper was to examine how well pretrained language models such as BERT and RoBERTa represent and process full scales of scalar adverbs in the absence of any specific task fine-tuning. We used naturalistic data from Reddit and also constructed sentences in order to explore the language models' ability to predict different types of scalar adverbs in context, and to distinguish them from negation. The models achieved some success when a left context of up to 40 words was available. However, we note many shortcomings: weak differentiation amongst the semantic classes of adverbs, poor ability to discriminate scalar adverbs from \textit{negations} even in contexts where a negation would create a contradiction, strong effects of adverb frequencies and lack of generalisation across two logically equivalent entailment constructions.




\section{Limitations}
\textbf{Scale:}
While our list of adverbs was carefully curated to include different semantic categories, full scales (including negations) and downsizer adverbs (e.g., \textit{slightly}) unlike in previous works, they are a restricted sample of only 24 adverbs. While we do believe this is a representative list, it is by no means an exhaustive one and the conclusions drawn in this paper have to be confined to the semantic categories explored. Thus we cannot exclude the possibility that experiments using a larger list of adverbs would produce different results.
\medskip

\noindent \textbf{Acceptability of negation}: For some of our natural stimuli, the use of \textit{not} would be infelicitous or illogical. For others, \textit{not} would be possible but would express a meaning that contradicts what the speaker chose to express. Without a large-scale exploration of alternative contextualizations of the items, it is difficult to separate these cases. 

For example, for item 498, a substitution of \textit{not} would appear to be rather infelicitous; however, by imagining further context we can see that this substitution would not be impossible.
\begin{enumerate}
    \item Including the bill itself, I think you'll be looking at around 600 A4 pages of reading.  And a lot of it is really dry.
    \item Including the bill itself, I think you'll be looking at around 600 A4 pages of reading.  And a lot of it is not dry.
    \item Including the bill itself, I think you'll be looking at around 600 A4 pages of reading.  And a lot of it is not dry. Your week might be more thrilling than you expect.
\end{enumerate}

\noindent\textbf{Models:}
While we tried to explore the predictions from different types of pretrained models (i.e., GPT and BERT), we acknowledge that we did not run an extensive study of models from different families. This is in part because these are the most commonly used models in applications, but also because our study is qualitative and we were mostly interested in comparing the models’ outputs with and without context and comparing performance between our semantic categories, rather than between different models. We also wished to focus on open-source models for which we could extract embeddings to explore a potential subspace for scalar properties.
\medskip

\noindent\textbf{Gold standard:}
There are few resources for providing gold standard labels of position on the scale for scalar adverbs in general, and especially when including different semantic categories and downsisers as well as maximisers. This limited our available choices for scalar adverbs to investigate. We provided the gold standard labels for the list of adverbs, based on information in WordNet and claims in the research literature, excluding adverbs whose semantics appeared unclear. While these rankings are informed by our best knowledge of semantics as experienced linguists, they were provided by a few researchers rather than by gathering judgments from many crowdsourcers as in other studies.

\section {Acknowledgements}

This study was supported by EPSRC Grant EP/W037211/1. We are grateful to Valentin Hofmann, Fangru Lin, and Paul R\"ottger for useful comments on a previous draft. 

\newpage

\bibliography{custom}
\bibliographystyle{acl_natbib}

\onecolumn
\newpage
\appendix
\section{Scalar ranking}

\begin{table}[htp]
\centering
\begin{tabular}{|l|l|l|l|l|l|l|l|}
\hline
& \textbf{Pacc.} 
& \multicolumn{3}{c|}{\textbf{Spearman $\rho$}} 
& \multicolumn{3}{c|}{\textbf{Kendall $\tau$}} \\
\hline
 
\textbf{\textit{BbS}} & 0.56  & 

\textit{f:} -0.28 &
\textit{m:} 0.58 &
\textit{d:} 0.68
& 
\textit{f:} -0.14 &
\textit{m:} 0.55 &
\textit{d:} 0.52
\\
 \hline
 
\textbf{\textit{BbD}} & 0.60 &

\textit{f:} 0.14 &
\textit{m:} \textbf{0.96} &
\textit{d:} \textbf{0.68}
& 
\textit{f:} 0.04 &
\textit{m:} \textbf{0.88} &
\textit{d:} \textbf{0.52}
\\
 \hline
 
\textbf{\textit{BbA}} &  0.60 &

\textit{f:} 0.68 &
\textit{m:} 0.77  &
\textit{d:} 0.32
& 
\textit{f:} 0.52 &
\textit{m:} 0.66 &
\textit{d:} 0.23
\\
\hline

\textbf{\textit{BlS}} &  0.57 & 

\textit{f:} -0.07 &
\textit{m:} 0.85 &
\textit{d:} 0.43
& 
\textit{f:}  -0.05 &
\textit{m:} 0.77 &
\textit{d: } 0.33 
\\
 \hline
 
\textbf{\textit{BlD}} & 0.54 & 

\textit{f:} -0.07 &
\textit{m:} 0.85 &
\textit{d:} 0.36
& 
\textit{f:} -0.05 &
\textit{m:}  0.77 &
\textit{d:} 0.33
\\
 \hline
 
\textbf{\textit{BlA}} & \textbf{0.64} & 

\textit{f:} \textbf{0.78} &
\textit{m:} 0.88 &
\textit{d:} 0.39
& 
\textit{f:} \textbf{0.62} &
\textit{m:} 0.77 &
\textit{d:} 0.24
\\
\hline

\textbf{\textit{RobS}} &  0.32 & 
\textit{f:} -0.82 &
\textit{m:}  -0.38 &
\textit{d:} 0.18
& 
\textit{f:} -0.62 &
\textit{m:}  -0.28 &
\textit{d: } 0.14
\\
 \hline
 
\textbf{\textit{RobD}} & 0.51 & 

\textit{f:} -0.46 &
\textit{m:}  0.81 &
\textit{d:} 0.54
& 
\textit{f:} -0.52 &
\textit{m:} 0.67 &
\textit{d:} 0.43
\\
 \hline
 
\textbf{\textit{RobA}} & 0.53 & 

\textit{f:} -0.32 &
\textit{m:} 0.77 &
\textit{d:} 0.64
& 
\textit{f:} -0.24 &
\textit{m:} 0.67 &
\textit{d:} 0.52
\\
\hline
\end{tabular}
\caption{\label{citation-guide}
Results of scalar ranking tests (\textbf{BbS} = BERT base SIM method, \textbf{BbD} = BERT base DIFF method, \textbf{BbA} = BERT base AdjDIFF method, \textbf{BlS} = BERT large SIM method, \textbf{BlD} = BERT large DIFF method, \textbf{BlA} = BERT large AdjDIFF method, \textbf{RobS} = RoBERTa SIM method, \textbf{RobD} = RoBERTa DIFF method, \textbf{RobA} = RoBERTa AdjDIFF method \textit{f} = FREQUENCY, \textit{m} = MODALITY. \textit{d} = DEGREE). 
}
\label{table:5}
\end{table}

\newpage
\section{Heatmaps}

\begin{figure}[ht]
        \centering
        \begin{subfigure}[b]{0.475\textwidth}
            \centering
            \includegraphics[width=\textwidth]{img/bertlarge_context.png}
            \caption[]%
            {{\small BERT large with context}}    
            \label{fig:mean and std of net14}
        \end{subfigure}
        \hfill
        \begin{subfigure}[b]{0.475\textwidth}  
            \centering 
            \includegraphics[width=\textwidth]{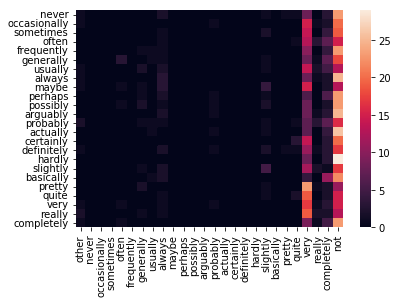}
            \caption[]%
            {{\small BERT large neutral context}}    
            \label{fig:mean and std of net24}
        \end{subfigure}
        \vskip\baselineskip
        \begin{subfigure}[b]{0.475\textwidth}   
            \centering 
            \includegraphics[width=\textwidth]{img/roberta_context.png}
            \caption[]%
            {{\small RoBERTa large with context}}    
            \label{fig:mean and std of net34}
        \end{subfigure}
        \hfill
        \begin{subfigure}[b]{0.475\textwidth}   
            \centering 
            \includegraphics[width=\textwidth]{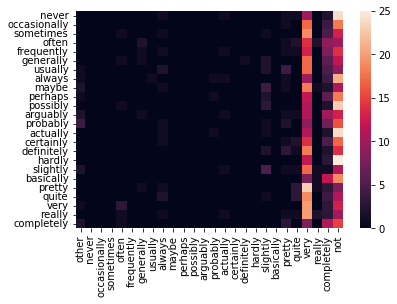}
            \caption[]%
            {{\small RoBERTa large neutral context}}    
            \label{fig:mean and std of net44}
        \end{subfigure}
        \caption[]
        {\small Heatmap of confusion matrices per intensifier for model and context in the MLM task (items are grouped by semantic category). In the interest of space considerations, we only show results for BERT large and RoBERTa.} 
        \label{fig:heatmaps_accuracies}
    \end{figure}

\begin{figure}[ht]
        \centering
        \begin{subfigure}[b]{0.475\textwidth}
            \centering
            \includegraphics[width=\textwidth]{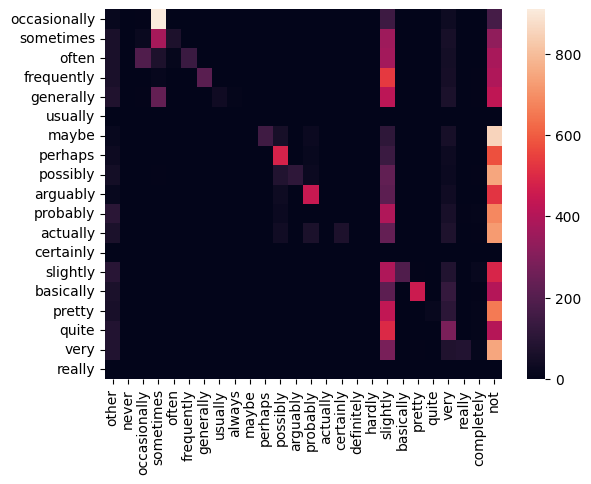}
            \caption[]%
            {{\small BERT large BELOW condition}}    
            \label{fig:mean and std of net14}
        \end{subfigure}
        \hfill
        \begin{subfigure}[b]{0.475\textwidth}  
            \centering 
            \includegraphics[width=\textwidth]{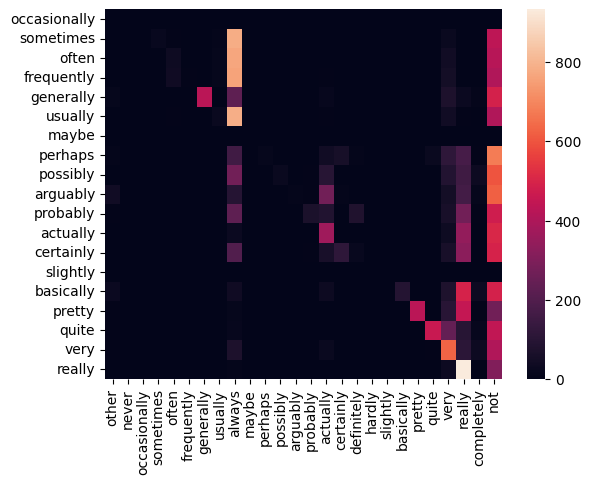}
            \caption[]%
            {{\small BERT large ABOVE condition }}    
            \label{fig:mean and std of net24}
        \end{subfigure}
        \vskip\baselineskip
        \begin{subfigure}[b]{0.475\textwidth}   
            \centering 
            \includegraphics[width=\textwidth]{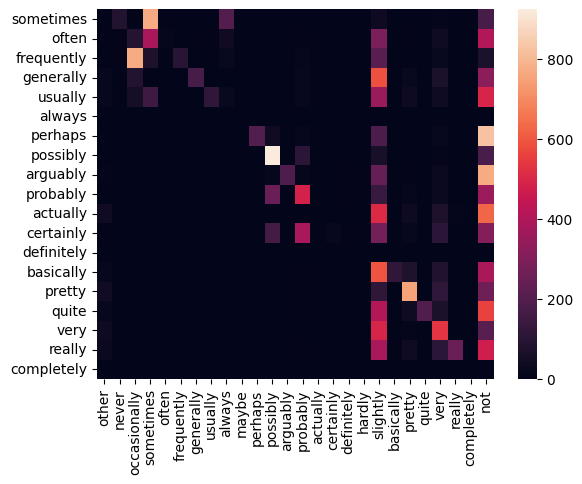}
            \caption[]%
            {{\small RoBERTa large BELOW condition}}    
            \label{fig:mean and std of net34}
        \end{subfigure}
        \hfill
        \begin{subfigure}[b]{0.475\textwidth}   
            \centering 
            \includegraphics[width=\textwidth]{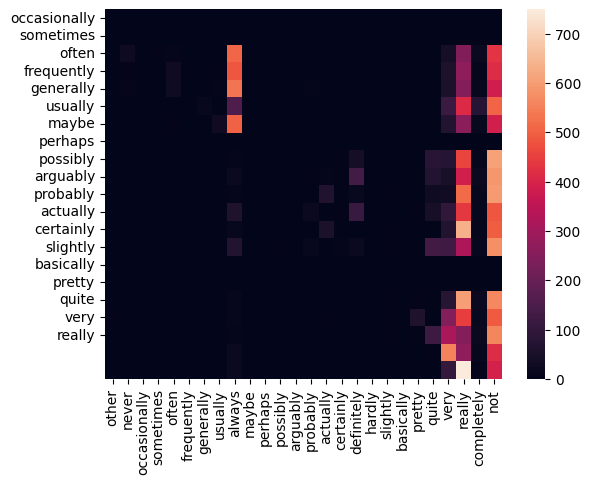}
            \caption[]%
            {{\small RoBERTa large ABOVE condition}}    
            \label{fig:mean and std of net44}
        \end{subfigure}
        \caption[]
        {\small Heatmap of confusion matrices for BELOW and ABOVE conditions including negations as answer. } 
        \label{fig:heatmaps_with_not}
    \end{figure}

\begin{figure}[ht]
        \centering
        \begin{subfigure}[b]{0.475\textwidth}
            \centering
            \includegraphics[width=\textwidth]{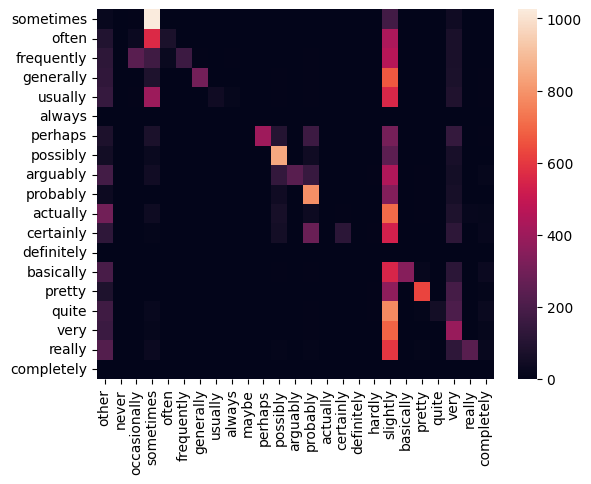}
            \caption[]%
            {{\small BERT large BELOW condition}}    
            \label{fig:mean and std of net14}
        \end{subfigure}
        \hfill
        \begin{subfigure}[b]{0.475\textwidth}  
            \centering 
            \includegraphics[width=\textwidth]{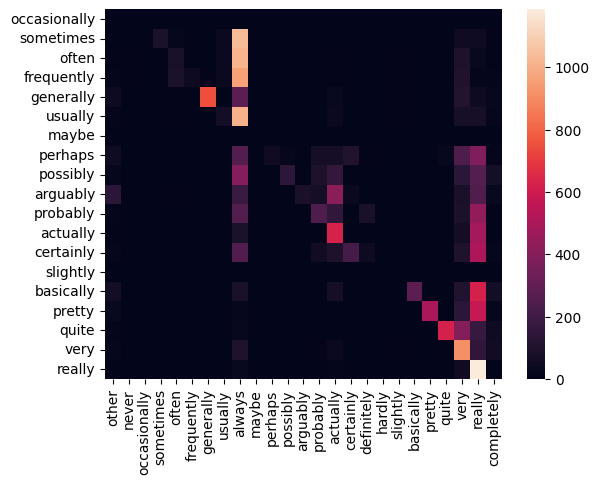}
            \caption[]%
            {{\small BERT large ABOVE condition }}    
            \label{fig:mean and std of net24}
        \end{subfigure}
        \vskip\baselineskip
        \begin{subfigure}[b]{0.475\textwidth}   
            \centering 
            \includegraphics[width=\textwidth]{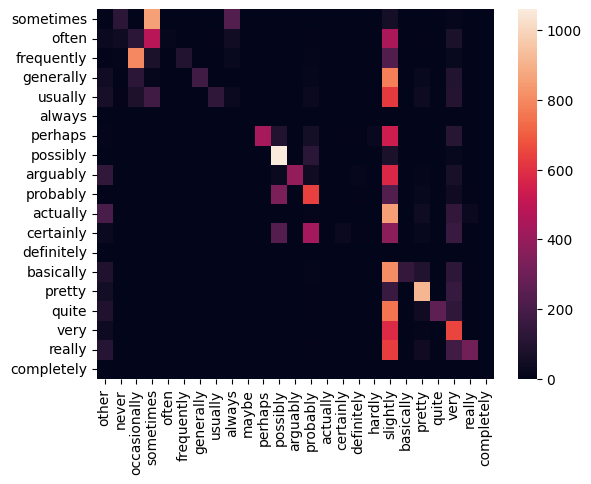}
            \caption[]%
            {{\small RoBERTa large BELOW condition}}    
            \label{fig:mean and std of net34}
        \end{subfigure}
        \hfill
        \begin{subfigure}[b]{0.475\textwidth}   
            \centering 
            \includegraphics[width=\textwidth]{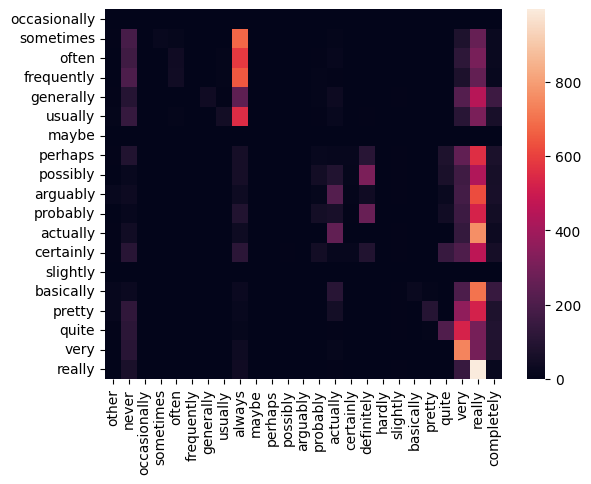}
            \caption[]%
            {{\small RoBERTa large ABOVE condition}}    
            \label{fig:mean and std of net44}
        \end{subfigure}
        \caption[]
        {\small Heatmap of confusion matrices for BELOW and ABOVE conditions of the entailment task, excluding negations as answer. } 
        \label{fig:heatmaps_no_not}
    \end{figure}

\clearpage
\section{Adverb frequencies}
\begin{table}[ht]
\begin{tabular}{|m{2.2cm}|m{4cm}|m{4cm}|} 
 \hline
 \textbf{ADVERB} & \textbf{Wordfreq} & \textbf{Reddit politosphere 2015} \\ 
 \hline
not & 0.34 & 0.44 \\
 never & 0.007 & 0.07\\
sometimes & 0.002 & 0.017\\
usually & 0.003  & 0.012\\
generally & 0.005  & 0.008\\
often & 0.004 & 0.023 \\
already & 0.01 & 0.03\\
frequently & 0.0004  & 0.003 \\
always & 0.02 & 0.05\\
maybe & 0.001 & 0.025 \\
perhaps & 0.0006 & 0.011\\
possibly & 0.002 & 0.005\\
probably & 0.02 & 0.02 \\
really & 0.078 & 0.08\\
actually & 0.03 & 0.027\\
certainly & 0.006 & 0.009\\
definitely & 0.007 & 0.008\\
slightly & 0.006 & 0.005  \\
hardly & 0.005 & 0.003 \\
basically & 0.004 & 0.004 \\
quite & 0.05 & 0.017 \\
pretty & 0.13 & 0.021 \\
very & 0.19 & 0.09 \\
seriously & 0.002 & 0.007 \\
completely & 0.07 & 0.009 \\
\hline
\end{tabular}
\caption{Relative frequencies of our scalar adverbs}
\label{table:freq}
\end{table}

\newpage

\section{Detailed results entailment task}

\begin{table}[ht]
\centering
\begin{tabular}{|c|m{1.2cm}|m{1.2cm}|m{1.6cm}|m{1.6cm}|} 
 \hline
 & \textbf{BERTl (acc.)} & \textbf{BERTl (triv.)}  
  & \textbf{RoBERTa (acc.)} & \textbf{RoBERTa (triv.)} \\ 
\hline
\textit{pseudo (\textit{below})} & 0.22  & 0.19   & 0.47 & 0.31 \\ 
\hline
 \textit{low (\textit{below})} &  0.30 & 0.19   & 0.38 & 0.23  \\
\hline
\textit{medium (\textit{below})} & 0.24  & 0.18   & 0.34 & 0.22 \\
\hline
\textit{high (\textit{below})} & 0.20  & 0.21   & 0.30 & 0.24 \\
\hline
\hline
\textit{pseudo (\textit{above})} & 0.39  & 0.23   & 0.45 & 0.15  \\
\hline
 \textit{low (\textit{above})} & 0.46  & 0.19   & 0.48 & 0.09 \\
\hline
\textit{medium (\textit{above})} & 0.46   & 0.19   & 0.49 & 0.09 \\
\hline
\textit{high (\textit{above})} & 0.49  & 0.22   & 0.51 & 0.10 \\
\hline
\hline

\textit{pseudo (\textit{below -no neg})} & 0.83  & 0.30   & 0.90 & 0.37  \\ 
\hline
 \textit{low (\textit{below -no neg)}} & 0.90  & 0.31   & 0.87 & 0.33 \\
\hline
\textit{medium (\textit{below -no neg)})} & 0.84  & 0.30   & 0.82 & 0.32 \\
\hline
\textit{high (\textit{below -no neg})} & 0.74  & 0.34   & 0.70 & 0.33 \\
\hline
\hline
\textit{pseudo (\textit{above -no neg})} & 0.95  & 0.40   & 0.96 & 0.24 \\
\hline
 \textit{low (\textit{above -no neg})}& 0.92  & 0.36   & 0.95 & 0.15  \\
\hline
\textit{medium (\textit{above -no neg})}&  0.92 & 0.32   & 0.95 &  0.15 \\
\hline
\textit{high (\textit{above -no neg})} & 0.92  & 0.32   & 0.95 & 0.15  \\
\hline

\end{tabular}
\caption{Results for scalar entailment dataset (BERT large and RobERTa). \textit{pseudo} = sentences with pseudo words from \citep{needle20224} \textit{low}= sentences with adjectives log frequencies -18 to -14, \textit{med}= sentences with adjectives log frequencies -14 to -10, \textit{high}= sentences with adjective log frequencies -10 to -6, \textit{(no neg)} = not taking into account negations as answers, \textit{acc.} = accuracy, \textit{triv.} = number of trivial answers (e.g., \textit{If it's sometimes ADJ, it's sometimes ADJ}, which we do not take into account for accuracies), \textit{below} = expects item below on the relevant scale, \textit{above} = expects item above on the relevant scale (best in bold).}
\label{entailresultsdetailed}
\end{table}

\section{Further probes}
\subsection{GPT-3}

GPT-3 is widely acknowledged to perform better than GPT-2. We probe GPT-3 (\textit{text-davinci-002}) \cite{DBLP:journals/corr/abs-2005-14165} on a sample of 5120 sentences from our dataset (from templates 1\_below and 1\_above given that the ABOVE condition for that particular template is the most difficult for RoBERTa), with a prompt mimicking the MLM task to obtain ten completions for the MASK token in our dataset. The results can be seen in Table \ref{entailresultsgpt} along with the results from RoBERTa on the same sample dataset. While GPT-3 outputs a much lower number of trivial and negative completions, it seems to be even more biased towards high frequency top-of-the-scale answers, causing it to perform poorly, especially in the BELOW condition. 

\begin{table}[htp]
\centering
\begin{tabular}{|c|m{1.2cm}|m{1.2cm}|m{1.6cm}|m{1.6cm}|} 
 \hline
 & \textbf{GPT-3 (acc.)} & \textbf{GPT-3 (triv.)}  
  & \textbf{RoBERTa (acc.)} & \textbf{RoBERTa (triv.)} \\

\hline
\textit{pseudo (\textit{below})} & 0.39  & 0.11   & 0.73  & 0.28  \\ 
\hline
 \textit{low (\textit{below})} & 0.41 & 0.09   & 0.6  & 0.20   \\
\hline
\textit{medium (\textit{below})} & 0.38  & 0.08    & 0.54  & 0.22  \\
\hline
\textit{high (\textit{below})} &  0.42 & 0.09   & 0.47  & 0.27  \\
\hline
\hline
\textit{pseudo (\textit{above})} & 0.40  & 0.13   & 0.0  & 0.0   \\
\hline
 \textit{low (\textit{above})} & 0.46  & 0.13  & 0.0  & 0.0  \\
\hline
\textit{medium (\textit{above})} & 0.51   & 0.08   & 0.0  & 0.0  \\
\hline
\textit{high (\textit{above})} &  0.47 & 0.06   & 0.0  & 0.0  \\
\hline
\hline
\textit{pseudo (\textit{below -no neg})} &  0.44 & 0.11   & 0.83  & 0.28  \\ 
\hline
 \textit{low (\textit{below -no neg})} & 0.47 & 0.09   & 0.79  & 0.21   \\
\hline
\textit{medium (\textit{below -no neg})} & 0.46  & 0.08    & 0.73  &  0.24 \\
\hline
\textit{high (\textit{below -no neg})} & 0.45  & 0.09   & 0.58  & 0.28  \\
\hline
\hline
\textit{pseudo (\textit{above -no neg})} &  0.70 & 0.13   & 1.0  & 0.13   \\
\hline
 \textit{low (\textit{above -no neg})} & 0.80  & 0.14  & 0.99  & 0.04  \\
\hline
\textit{medium (\textit{above -no neg})} &  0.89  & 0.09   & 0.97  & 0.08  \\
\hline
\textit{high (\textit{above -no neg})} & 0.81 & 0.06    & 0.97  & 0.08  \\
\hline
\hline
\hline
\textbf{with negation} & 0.43 & 0.10    & 0.29 & 0.12    \\
\hline\textbf{without negation} & \textbf{0.62}  & \textbf{0.10}   & \textbf{0.85}  & \textbf{0.18} \\
\hline
\hline
\textbf{BELOW} & 0.43 & 0.09    & \textbf{0.75}  & \textbf{0.25}   \\
\hline
\textbf{ABOVE} & \textbf{0.63} &  \textbf{0.10} & 0.49 & 0.04  \\
\hline

\end{tabular}
\caption{Results for sample scalar entailment dataset (GPT-3 and RoBERTa). \textit{pseudo} = sentences with pseudo words from \citep{needle20224} \textit{low} = sentences with adjectives log frequencies -18 to -14, \textit{med} = sentences with adjectives log frequencies -14 to -10, \textit{high} = sentences with adjective log frequencies -10 to -6, \textit{(no neg)} = not taking into account negations as answers, \textit{acc.} = accuracy, \textit{triv.} = number of trivial answers (e.g., \textit{If it's sometimes ADJ, it's sometimes ADJ}, which we do not take into account for accuracies), \textit{below} = expects item below on the relevant scale, \textit{above} = expects item above on the relevant scale (best in bold).}
\label{entailresultsgpt}
\end{table}

\subsection{NLI trained model}

To evaluate whether models which have been fine-tuned on a Natural Language Inference dataset (e.g., MNLI, \citealp{N18-1101}) we also adapt the dataset to the MNLI format.   We adapted the items by replacing the [MASK] token at random with an adverb that created a correct entailment for half our items and an incorrect one for the remaining items (e.g., \textit{It is always cold. It is sometimes cold.} vs. \textit{It is sometimes cold. It is always cold.'}. We obtain predictions from MNLI fine-tuned, BERT large and RoBERTa models. 
The results can be found in Table \ref{entailresultsnli}. The models perform near chance, indicating that training on a dataset containing broadly defined inferences such as MNLI does not improve performance on a strict entailment task involving scalar adverbs. 

\begin{table}[htp]
\centering
\begin{tabular}{|c|m{1cm}|m{1.5cm}|} 
 \hline
 & \textbf{BERTl (acc.)}   
  & \textbf{RoBERTa (acc.)} \\

\hline
 \textit{\textit{pseudo}} & 0.46   &   0.45  \\
\hline
\textit{\textit{low}} & 0.46  &   0.44  \\
\hline
\textit{\textit{med}} & 0.45  &   0.42  \\
\hline
\textit{\textit{high}} &  0.43 & 0.41   \\ 
\hline
\hline
\textbf{\textit{avg.}} &  \textbf{0.45} & \textbf{0.43}   \\ 
\hline

\end{tabular}
\caption{Results for scalar entailment with models fine tuned on MNLI \cite{N18-1101}. \textit{pseudo} = sentences with pseudo words from \citep{needle20224} \textit{low} = sentences with adjectives log frequencies -18 to -14, \textit{med} = sentences with adjectives log frequencies -14 to -10, \textit{high} = sentences with adjective log frequencies -10 to -6, \textit{acc.} = accuracy (dataset is balanced with chance = 0.50)}
\label{entailresultsnli}
\end{table}

\newpage

\section{Dataset characteristics}

\begin{table}[htp]
\centering
\begin{tabular}{|c||m{1.5cm}|m{1cm}|m{1cm}|m{1cm}|m{1cm}|m{2cm}|} 
 \hline
 \textbf{category}   & \textbf{samples} & \textbf{min words} & \textbf{max words} & \textbf{avg words} & \textbf{n\_adj} & \textbf{adj. overlap} \\ 
\hline
\hline
 \textit{frequency} & 320  & 4 & 59 & 22.08 & 273 & 1.92 \\
\hline
\textit{modality} & 320 & 4 & 53 & 21.97 &  283 & 1.57\\
\hline
\textit{degree} & 320 & 4 & 48 & 21.33 &  304 & 0.57\\
\hline

\end{tabular}
\caption{word length and adjective characteristics of dataset: \textit{samples}=n samples in category; \textit{min words}=minimum word length for each sample; \textit{max words}=maximum word length for each sample; \textit{n\_adj}=number of adjectives; \textit{adj overlap}=average adjective overlap between adverbs}
\label{naturalisticdatasetchar}
\end{table}

\newpage

\section{Entailment results with templates}
\begin{table}[htp]
\centering
\begin{tabular}{|m{5cm}|m{1.5cm}|m{1cm}|m{1cm}|m{1.5cm}|m{1.5cm}|} 
 \hline
\textbf{Example template} & \textbf{Id} & \textbf{BERTl (acc.)} & \textbf{BERTl (triv.)}  
  & \textbf{RoBERTa (acc.)} & \textbf{RoBERTa (triv.)} \\

\hline
\textit{If it is often cold, then it is at least [MASK] cold. \textit{(below)}} & 1\_below & 0.51  & 0.58   &  0.59  & 0.24  \\ 
\hline
 \textit{It is at least [MASK] cold if it is often cold. (\textit{below})} & 2\_below & 0.12  & 0.16    & 0.22  & 0.11   \\
\hline
\textit{It is often cold so it is at least [MASK] cold. (\textit{below})} & 3\_below & 0.32  & 0.18    & 0.41  & 0.33  \\
\hline
\textit{It is at least [MASK] cold because it is often cold. (\textit{below})} & 4\_below &  0.05 & 0.08  & 0.38  & 0.56  \\
\hline
\textit{If it is at most [MASK] cold, then it is not often cold. (\textit{below})} & 5\_below & 0.63  & 0.26   & 0.37  & 0.39  \\ 
\hline
 \textit{It is not often cold if it is at most [MASK] cold. (\textit{below})} & 6\_below & 0.24  & 0.07   & 0.38  & 0.11   \\
\hline
\textit{It is at most [MASK] cold so it is not often cold. (\textit{below})} & 7\_below & 0.24   & 0.15    & 0.32  & 0.15  \\
\hline
\textit{It is not often cold because it is at most [MASK] cold. (\textit{below})} & 8\_below &  0.12  & 0.09    & 0.37   & 0.09   \\

\hline
\hline
\textit{If it is [MASK] blue, then it is at least quite blue. (\textit{above})} & 1\_above & 0.01 & 0.04  & 0.0  & 0.0   \\
\hline
\textit{It is at least quite blue if it is [MASK] blue. (\textit{above})} & 2\_above & 0.03 & 0.07   & 0.01  & 0.02   \\
\hline
\textit{It is [MASK] blue so it is at least quite blue. (\textit{above})} & 3\_above &0.15 & 0.10   & 0.25  & 0.16   \\
\hline
\textit{It is at least quite blue because it is [MASK] blue. (\textit{above})} & 4\_above & 0.42 & 0.26   & 0.22  & 0.09   \\
\hline
\textit{If it is at most quite blue, then it is not [MASK] blue.  (\textit{above})} & 5\_above & 0.93 & 0.41   & 0.92  & 0.18   \\
\hline
\textit{It is not [MASK] blue if it is at most quite blue.  (\textit{above})} & 6\_above & 0.87 & 0.27  & 0.96  & 0.10   \\
\hline
\textit{It is at most quite blue so it is not [MASK] blue. (\textit{above})} & 7\_above & 0.81  & 0.25   & 0.91  & 0.15   \\
\hline
\textit{It not quite blue because it is at most [MASK] blue. (\textit{above})} & 8\_above & 0.12  & 0.09    & 0.37   & 0.09    \\

\hline
\hline
\textbf{MASK before premise} & NA & 0.29  &  0.14  &  0.36  &  0.20  \\
\hline
\textbf{premise before MASK} & NA &\textbf{0.39}  & \textbf{0.22}   & \textbf{0.46}    & \textbf{0.12}    \\

\hline

\end{tabular}
\caption{Results for scalar entailment per template and example template (with negations) (best in bold).}
\label{entailresultstemplatesneg}
\end{table}

\begin{table}[htp]
\centering
\begin{tabular}{|m{5cm}|m{1.5cm}|m{1cm}|m{1cm}|m{1.5cm}|m{1.5cm}|} 
 \hline
\textbf{Example template} & \textbf{Id} & \textbf{BERTl (acc.)} & \textbf{BERTl (triv.)}  
  & \textbf{RoBERTa (acc.)} & \textbf{RoBERTa (triv.)} \\

\hline
\textit{If it is often cold, then it is at least [MASK] cold. \textit{(below)}}  & 1\_below & 0.80  & 0.59   & 0.74  & 0.26  \\ 
\hline
 \textit{It is at least [MASK] cold if it is often cold. (\textit{below})} & 2\_below & 0.89  & 0.43    & 0.84  & 0.32   \\
\hline
\textit{It is often cold so it is at least [MASK] cold. (\textit{below})} & 3\_below & 0.80  & 0.26    & 0.76  & 0.38  \\
\hline
\textit{It is at least [MASK] cold because it is often cold. (\textit{below})} & 4\_below & 0.85  & 0.32   & 0.86  & 0.65  \\
\hline
\textit{If it is at most [MASK] cold, then it is not often cold. (\textit{below})} & 5\_below & 0.97  & 0.31   & 0.97  & 0.48  \\ 
\hline
 \textit{It is not often cold if it is at most [MASK] cold. (\textit{below})} & 6\_below & 0.80  & 0.16    & 0.85  & 0.15   \\
\hline
\textit{It is at most [MASK] cold so it is not often cold. (\textit{below})} & 7\_below &0.79  & 0.24    & 0.82  & 0.27  \\
\hline\textit{It is not often cold because it is at most [MASK] cold. (\textit{below})} & 8\_below & 0.92   &  0.27  &  0.98 & 0.17   \\

\hline
\hline
\textit{If it is [MASK] blue, then it is at least quite blue. (\textit{above})} & 1\_above & 0.92 & 0.50  & 0.98  & 0.08   \\
\hline
\textit{It is at least quite blue if it is [MASK] blue. (\textit{above})} & 2\_above & 0.90 & 0.28   & 0.92  & 0.16   \\
\hline
\textit{It is [MASK] blue so it is at least quite blue. (\textit{above})} & 3\_above & 0.86 & 0.42  & 0.87  & 0.30   \\
\hline
\textit{It is at least quite blue because it is [MASK] blue. (\textit{above})} & 4\_above & 0.95 & 0.37  & 0.96  & 0.24   \\
\hline
\textit{If it is at most quite blue, then it is not [MASK] blue.  (\textit{above})} & 5\_above & 0.99 & 0.42   & 0.93  & 0.18   \\
\hline
\textit{It is not [MASK] blue if it is at most quite blue.  (\textit{above})} & 6\_above & 0.93  & 0.28   & 0.99  & 0.10   \\
\hline
\textit{It is at most quite blue so it is not [MASK] blue. (\textit{above})} & 7\_above & 0.94 & 0.26   & 0.96  & 0.16   \\
\hline
\textit{It not quite blue because it is at most [MASK] blue. (\textit{above})} & 8\_above & 0.67  & 0.20    & 0.76   & 0.17    \\
\hline
\hline
\textbf{MASK before premise} & NA & \textbf{0.88}  & \textbf{0.38}    & \textbf{0.90}    & \textbf{0.31}    \\
\hline
\textbf{premise before MASK} & NA & 0.86 &   0.31  &  0.87  & 0.20   \\
\hline

\end{tabular}
\caption{Results for scalar entailment per template and example template (without negations) (best in bold).}
\label{entailresultstemplates}
\end{table}

\end{document}